\DocumentMetadata{}
\documentclass[sigconf]{acmart}

\usepackage{dirtytalk}
\usepackage{xspace}
\renewcommand{\paragraph}[1]{\medskip\noindent\textbf{#1}}
\newcommand{\etal}{\textit{et al.}\xspace}

\copyrightyear{2025}
\acmYear{2025}
\setcopyright{rightsretained}
\acmConference[KDD '25] {Proceedings of the 31st ACM SIGKDD Conference on Knowledge Discovery and Data Mining V.2}{August 3--7, 2025}{Toronto, ON, Canada.}
\acmBooktitle{Proceedings of the 31st ACM SIGKDD Conference on Knowledge Discovery and Data Mining V.2 (KDD '25), August 3--7, 2025, Toronto, ON, Canada}
\acmDOI{10.1145/3711896.3736562}
\acmISBN{979-8-4007-1454-2/2025/08}
\begin{document}

\title{Synthetic Tabular Data: Methods, Attacks and Defenses}

\author{Graham Cormode}
\affiliation{
\institution{Meta and University of Warwick}  
\city{Coventry} 
\country{UK}}
\email{gcormode@meta.com}

\author{Samuel Maddock}
\affiliation{
\institution{Meta and University of Warwick} 
\city{Menlo Park} 
\country{USA}}
\email{smaddock@meta.com}

\author{Enayat Ullah}
\affiliation{
\institution{Meta} 
\city{Menlo Park} 
\country{USA}}
\email{enayat@meta.com}

\author{Shripad Gade}
\affiliation{
\institution{Meta} \city{Menlo Park} 
\country{USA}}
\email{shripadgade@meta.com}

\begin{abstract}
Synthetic data is often positioned as a solution to replace sensitive fixed-size datasets with a source of  unlimited matching data, freed from privacy concerns. 
There has been much progress in synthetic data generation over the last decade, leveraging corresponding advances in machine learning and data analytics. 
In this survey, we cover the key developments and the main concepts in tabular synthetic data generation, including paradigms based on probabilistic graphical models and on deep learning. 
We provide background and motivation, before giving a technical deep-dive into the methodologies. 
We also address the limitations of synthetic data, by studying attacks that seek to retrieve information about the original sensitive data. 
Finally, we present extensions and open problems in this area. 
\end{abstract}



\maketitle

\section{Introduction}

A common scenario in many data-focused applications is when there is a valuable dataset but its contents are very sensitive. 
For instance, this could be a dataset of customers with their personal details and purchases, or a dataset of hospital patients with information on their health conditions. 
The dataset would be very useful to share with data scientists or machine learning engineers, but due to privacy concerns it is not appropriate to make the data available in its original form.  Instead we would like to create a new dataset that shares the characteristics of the original data, but is entirely fabricated. 
This is referred to as ``Synthetic Data Generation''.  
Being completely made up, intuitively we would believe synthetic data is freed of privacy concerns, and can be shared more easily than the original source. 
However, things are not so simple: if the synthetic data is very similar to the original data, it may leak sensitive information about its source.  
Meanwhile, if the synthetic data does not resemble the original data, it is not a useful substitute. 
Research in synthetic data is concerned with walking this tightrope: balancing fidelity and privacy, whilst taking into account expressivity (richness of the model), and efficiency (computational cost).  

Synthetic data can take many forms, depending on the domain.  
We might want to generate synthetic text, synthetic images and videos, or synthetic three-dimensional objects. 
However, in this survey we focus on the core case of \textit{synthetic tabular data}: data which is most naturally represented within a structured table. 
This captures many problems in data management, where we can consider the tables as relations from a database; and in machine learning, where the rows are examples and the columns are features.

This survey aims to give an overview of the state-of-the-art in synthetic tabular data. 
We will describe the objectives and desiderata for synthetic data, and how they are achieved. 
We will show how techniques have developed from simplistic modeling to leveraging complex cutting-edge machine learning models, and the tradeoffs along this path. 
A number of different lenses can be used to view the task of generating synthetic data: 
a statistical lens, which seeks to find a parsimonious model of the original data from which new examples can be sampled (Section~\ref{sec:marginals}); or a machine learning perspective, which seeks to train a model that can generate examples that are sufficiently realistic to fool a classifier; we can also adopt the framing of generative AI, where the objective is to create data based on many real-world examples of tables and the context of a specific target (Section~\ref{sec:dl}). 
We will also consider the limitations of synthetic data generation, and how adversaries can try to use the output synthetic data to learn private information about the data that the model was trained on (Section~\ref{sec:mia}).  
We will address defenses against such attacks based on formal privacy guarantees, and discuss how the relative success of attacks can be used as an estimate of empirical privacy risk of synthetic data release. 
We conclude with a consideration of extensions to other forms of data and other scenarios, and open problems for the community to work on (Section~\ref{sec:open}). 

\section{Background and Preliminaries}

\subsection{Social and Legal Reasons}

The requirements for privacy stem from many sources. 
Legal frameworks such as GDPR and CCPA place restrictions on what data can be shared and under what protections. 
Organizations handling data make promises to data subjects about how their information will be processed and protected. 
Consumers exercise their freedom of choice, and may prefer to use one organization's services over another based on their privacy stance. 
Collectively, these mean that data must be handled with care, and appropriate controls put in place to restrict how it is used. 
At the same time, many organizations \textbf{utilize} sensitive data, \textbf{for example} to inform their business decisions.  
To bridge this gap, various privacy enhancing technologies (PETs) have been proposed in order to provide the results of computations over sensitive data while offering suitable privacy guarantees on what is not revealed. 

Synthetic data generation is a popular privacy enhancing technology, but the methodologies to model data and generate synthetic datasets is still evolving. 
Except in trivial cases, synthetic data requires a \textit{reference dataset} to work from. 
That is, unless it is satisfactory to use an off-the-shelf statistical distribution (e.g., a Normal distribution with mean zero and unit variance) or naive modeling (e.g., a table where each attribute is chosen independently and uniformly at random), we need to generate the synthetic data with some properties of the real dataset. 
It is therefore useful to  treat this as a machine learning problem: given some training data (the reference dataset), the task is to learn a model that can generate new examples, such that the generated data resembles the training data. 
The process must ensure that some privacy guarantees hold (to prevent trivially non-private approaches, such as copying the training data in whole or in part), and may additionally try to optimize some utility metrics.  
This immediately sets some requirements for synthetic data: there must be reference data available to guide the modeling. 
The applications must tolerate some imprecision, since the synthetic data will necessarily not be a perfect match for the training data, and the reference data must have enough examples to allow the synthetic data generation to learn patterns. 
This already restricts some scenarios, such as in healthcare, where if there are only a very small number of cases of a particular rare disease, changing any details risks disguising an important pattern.  
For such settings, other privacy enhancing technologies, such as the use of secure data environments, might be a better fit.

\subsection{Desiderata for Synthetic Data}

There are multiple desirable properties for synthetic data, as outlined above.  
\begin{itemize}
    \item \textbf{Fidelity:} the generated data must be ``faithful'' to the reference data.  This may be measured via an appropriate distance metric, or by comparing various statistical characteristics of the data e.g., mean/median values, frequency distributions, and correlations. 
    \item \textbf{Privacy:} the result of modeling the data, either in the form of the model itself or the data that it generates, must not leak sensitive information about the reference data under some appropriate formalization.  
    \item \textbf{Efficiency:} the process must be computationally efficient to perform.  
    It may be the case that synthetic generation is done only once for a given reference dataset and the results are used extensively, so the process does not have to be real-time, but still operate within reasonable bounds on time and memory requirements. 
    \item \textbf{Expressivity:} the model must be sufficiently rich to capture sufficient details of the reference data. 
    Typically, we want the generated data to match the format of the reference data: the same attributes and domain. 
    \item \textbf{Utility:}
    the data should be useful for downstream tasks, allowing similar conclusions to be drawn as with the original data. 
    This may be tested by application-specific measures, e.g., evaluating the performance of ML models trained on some (withheld) reference data. 
\end{itemize}

Although fidelity and utility may appear similar, they focus on different aspects of synthetic data generation. 
It may be possible to adopt generic measures of fidelity based on the statistical distance between the generated and reference data, whereas in general there is no universal definition of utility, and the same synthetic data may have very different utilities for different tasks.  
We will revisit each of these properties as we discuss different methodologies.  
Next, we provide more detail on how privacy can be formalized. 

\subsection{Privacy Requirements}

Defining ``privacy'' in formal terms has been a challenging problem for the computer science and statistical communities for many years. 
Initial attempts to give syntactic properties that correspond to privacy are seen to be lacking due to loopholes~\cite{Clifton:Tassa:13}. 
This is particularly the case for synthetic data, where syntactic requirements on the output can be easily evaded. 
For instance, the $k$-anonymity requirement states that each record in the output must match at least $k-1$ others \cite{sweeney2002k}. 
This can be defeated for synthetic data by simply duplicating each true record $k$ times. 

In this survey, we will focus on the statistical notion of Differential Privacy (DP), which has become the de facto standard \cite{Dwork:Roth:14}. 
The definition of differential privacy states (informally) that any property of the output should be approximately equally likely whether or not any individual was included in the input.  
This means the synthetic data (or the parameters of the model used to generate it) should not depend too strongly on the contribution of any one individual. 
There are many ways to achieve differential privacy for a given task, but it is most often achieved through the introduction of carefully calibrated random noise into the numeric values that are computed by the algorithm.  
There are also many variations of the definition, which affect the exact nature of the privacy guarantee and the mathematical properties that support the privacy analysis -- including local differential privacy (LDP)~\cite{kasiviswanathan2011can}, pure/approximate privacy~\cite{Dwork:Roth:14}, R\'enyi privacy (RDP)~\cite{Mironov:17}, and zero-concentrated differential privacy (zCDP)~\cite{Bun:Steinke:16}. 
For the purposes of this survey, we will largely gloss over the differences between these concepts for the presentation of different methods. 
For the privacy expert, it suffices to state that the majority of results in synthetic data generation are characterized in terms of approximate DP, although bounds are often proved via RDP or zCDP techniques. 

\section{Marginal-based methods}
\label{sec:marginals}
\subsection{Initial Statistical Approaches}

Initial approaches to synthetic data generation emerged from the statistical community in the late 20th century. 
These proceed by manipulating the reference data, for sources such national census data. 
The concept of ``data swapping''~\cite{Dalenius:Reiss:82} involves exchanging the attribute values for various individuals in the data.  
This approach preserves some properties exactly (such as the marginal distribution of each attribute), but risks introducing impossible combinations, as correlations may not be preserved. 
Subsequent work on this approach is surveyed by Fienberg and McIntyre~\cite{Fienberg:McIntyre:04}. 

A related example is SMOTE, Synthetic Minority Over-sampling technique~\cite{Bowyer:Chawla:Hall:Kegelmeyer:02}.
SMOTE was proposed in the context of handling data imbalance, where certain classes were under-represented, and so it was desirable to generate additional synthetic examples of those classes. 
Here, records are assumed to lie within in a geometric space. 
A synthetic point is generated by taking a point $p$ from the reference dataset, and picking one of its nearest neighbors $q$.  
The new point is formed by interpolating between $p$ and $q$, as $\alpha p + (1-\alpha)q$, for $\alpha$ chosen randomly in $[0,1]$. 
These approaches may be useful for generating realistic-looking examples, but may not give any useful privacy guarantee if there are several similar individuals in the data: swapping or interpolating their values will give near-duplicate examples. 

\subsection{Probabilistic Graphical Models}

Subsequent work on synthetic data takes inspiration from the statistical modeling, where the key concepts include  \textit{marginals} and \textit{probabilistic graphical models (PGMs)}~\cite{Koller:Friedman:09}. 
  In what follows, most works concentrate on the case of a single dataset in the form of a table with $n$ rows and $d$ columns. 
  Each row corresponds to an individual record, while each column corresponds to an attribute (or feature). 
  As a canonical example, consider a demographic dataset where each row refers   to an individual, and the attributes are their properties, such as age, sex, country of residence, income level etc. 
  As a further simplification, we will assume that each feature is treated as categorical, and so is drawn from a known set of possibilities. 

Given such a dataset $X$, we can define an (empirical) probability distribution based on the frequency of observed values. 
For instance, for the country of residence, we can define $\Pr_\text{country}$ as the distribution of countries seen in the data.  
Then, for instance, $\Pr_\text{country}[ \text{USA}] = | \{ i : X_{i,j} = \text{`USA'} \}| / n$, if $j$ is the column of $X$ containing country of residence information.  
More generally, we can similarly find probabilities for combinations of values, e.g., 
$\Pr_\text{country,sex}[\text{USA, female}]$.  
These collections of probabilities are referred to as \textit{marginal distributions}, or marginals for short, since they correspond to aggregations that could be computed at the margin of tables, after summing over some removed (or `marginalized') columns. 
The distribution over (country, sex) is referred to as a 2-way marginal, since it includes two attributes, and more generally we may refer to $k$-way marginals for various values of $k$. 

A probabilistic graphical model (or just `graphical model') is a way of representing the interaction of multiple variables in a compact way, possibly at the expense of some fidelity~\cite{Koller:Friedman:09}. 
In our setting, we will consider the full distribution defined by the reference data, also known as the \textit{joint distribution}, and consider approaches to capture this via a smaller collection of marginals. 
The area of probabilistic graphical models studies ways to do so effectively. 
A graphical model uses a graph structure to represent the structure: each node corresponds to an attribute, and edges link nodes that are correlated. 
Nodes without an edge directly linking them are considered to be (conditionally) independent, and so the relationship between the corresponding attributes is not explicitly described in the model. 
Attributes whose correlations are captured by the model have this correlation described by a marginal distribution.  

Several synthetic data generation methods can be understood as modeling the reference data via a graphical model. 
The graphical model is represented by the graph structure and a set of marginal distributions. 
The marginal distributions are learned from the reference data, and typically have some random noise added to provide a formal privacy guarantee.
The structure of the graphical model means that synthetic data can be sampled from this representation, sometimes referred to as `inference'.

\subsection{Early Work}

A first approach to creating synthetic data from marginals is to model just the correlations between each attribute and one ``target'' attribute in turn. 
This corresponds to the textbook Naive Bayes model~\cite{NaiveBayes}. 
In more detail, assume that we pick attribute $d$ as the ``target'' attribute. 
The Naive Bayes approach is to build the two-way marginal distributions between each attribute $j<d$ and $d$. 
Privacy can be achieved by adding suitable random noise to each marginal. 
Synthetic data is generated, for instance, by first sampling a value for the target attribute according its own (one-way) marginal distribution. 
Then each other attribute is filled in by sampling from the corresponding two-way marginal, conditioned on the value chosen for the target attribute. 
This approach is simple to instantiate, but fails to capture any of the complex structure of the underlying reference data. 

At the other extreme is the Multiplicative Weights with Exponential Mechanism (MWEM) approach~\cite{Hardt:Ligett:McSherry:12}. 
This approach maintains probabilities over the full joint distribution of $d$ attributes (so the size of the representation is exponential in $d$).  
The method also requires a collection of (linear) queries $Q$ that should be answered accurately: this defines a `workload'.  
It builds a private distribution $A$ over the attributes, initialized to be uniform. 
The essence of the method is to iteratively improve the distribution $A$ by picking (privately) a query $q \in Q$ that is currently poorly answered. 
That is, pick query $q$ as a function of $q(A) - q(X)$. 
Then $A$ is updated in order to improve the result for query $q$, following the multiplicative weights definition~\cite{Hardt:Rothblum:10}. 
Synthetic data can then be drawn by directly sampling from $A$. 
This method has strong privacy and utility guarantees, but won't scale well for even moderate values of $d$.  
  
\subsection{PrivBayes: a middle way} 
Naive Bayes and MWEM adopt two extreme approaches: considering only limited correlations or considering the full dataset. 
Naive Bayes misses information about combinations of attributes, while MWEM risks allowing the sparse distribution being drowned out by privacy noise. 
A middle way is to seek a representation that materializes information that is just rich enough to describe useful correlations without losing information in the noise. 
The PrivBayes approach adopts the model of Bayesian networks to represent the data~\cite{ZhangCPSX17}. 
A Bayesian network is specified by a graph $G$, whose nodes are the $d$ attributes in the data. 
Given $G$, it is straightforward to materialize the corresponding set of marginals, and add the noise required to satisfy differential privacy. 
Hence, most of the work for PrivBayes is in choosing what graph $G$ to use. 

PrivBayes includes a number of steps to choose $G$ given dataset $X$. 
It takes a greedy approach, considering a set of possible marginals based on an information theoretic measure of how much value they add.
Specifically, it considers the mutual information of a marginal, defined for a pair $X, Y$ as
\begin{equation}
I(X; Y) = \sum_{x,y}\Pr(x, y) (\log \Pr(x, y) - \log(\Pr(x)\Pr(y))). 
\label{eq:mutualinfo}
\end{equation}
Exact computation of mutual information entails adding a large amount of privacy noise, so PrivBayes replaces this with a surrogate measure that is more privacy-friendly. 
Attention is restricted to only the $k$-way marginals, where $k$ is chosen based on a heuristic combining $n$, $d$, and the differential privacy parameter $\varepsilon$. 
PrivBayes is relatively simple to instantiate and apply, and as a result it has enjoyed popularity as a baseline approach to private synthetic data generation. 
For instance, it has been adopted to publish demographic national statistics on live births~\cite{Hod:Canetti:24}.

\subsection{Richer Models}  
The PrivBayes approach provides a robust benchmark for synthetic data generation, but has many opportunities for extension and improvement.  
Successive approaches have expanded on the graphical modeling approach, by varying the class of models considered, and the approach taken to learn the model structure.  

The (Private-)PGM algorithm~\cite{McKennaSM19} addresses the general problem of estimating and sampling from a high-dimensional distribution (i.e., inference) given a collection of noisy measurements of a set of marginals. 
It adopts the idea of using a workload of (linear) queries to guide the synthetic data generation, similar to MWEM. 
It formalizes the task as one of optimization, to choose a set of queries to pose on the data, so that the answer to these queries is sufficient to instantiate a model for synthetic data generation and that the error of the query workload will be minimized. 
The optimization proceeds by minimizing a convex loss function, which produces the parameters of a graphical model as a by-product. 
The approach can be used to find solutions for a variety of graphical models, provided the graphical model is given as input, including the full joint distribution (i.e., the MWEM setting), Bayesian networks (i.e., the PrivBayes setting), and more. 
The PGM algorithm has been widely used subsequently as a subroutine in many synthetic data generation methods. 

The MST approach~\cite{McKennaMS21} makes use of the PGM algorithm and adapts the modeling approach of PrivBayes. 
Specifically, it builds a graph over the attributes where edge weights are defined based on the mutual information between the two corresponding attributes. 
It then computes the maximum spanning tree (MST) over this graph, and materializes the marginals corresponding to the edges of tree. 
This step is inspired by the Chow-Liu algorithm for Bayes net construction~\cite{Chow:Liu:68}. 
Then it considers three-way marginals that correspond to attributes that are neighbors of the tree edges, and a maximum spanning tree of the resulting graph is computed. 
Marginals from this tree are also materialized, and finally PGM is used to generate the data from the captured information. 

The PrivMRF algorithm starts from the observation that Bayesian networks inherently limit the expressivity of the model, and that Markov Random Fields (MRF) present a more general model~\cite{PrivMRF}. 
For a Markov Random Field, the graph $G$ is replaced with a hypergraph $H$ (so every Bayesian network is an MRF, but not vice-versa). 
Learning an MRF under privacy is more complicated, 
and PrivMRF seeks to select a set of marginals to materialize that accurately describe the reference dataset while ensuring that the inference step remains efficient. 
PrivMRF selects pairs of marginals to materialize when their joint distribution is very different from the product of each distribution separately.
Mathematically, this is 
\begin{equation} D(X, Y) = \sum_{x,y} \Pr(x,y) - \Pr(x)\Pr(y)\end{equation} 
for distribution pairs $X, Y$ --- note that this is similar in spirit, but different in detail from mutual information in~\eqref{eq:mutualinfo}. 
It then performs a triangulation of the resulting graph, ensuring that the cliques in the triangulation are not too large, since large cliques correspond to a computationally expensive inference task. 
PrivMRF uses the PGM algorithm as a key step to obtain the marginal distributions from the triangulated graph representation. 

The PrivSyn algorithm seeks to handle scenarios where there are many attributes, and the attributes have high cardinality~\cite{PrivSyn}. 
Rather than explicitly using a graphical model, PrivSyn picks a large collection of overlapping low-degree marginal distributions (one-way and two-way marginals). 
It picks marginals based on the same criterion as PrivMRF, choosing attribute pairs whose joint distribution is most different from their product distribution. 
The selected marginals are published with noise, then post-processing is applied to achieve consistency among the overlapping information. 
To produce synthetic data, an approach similar to MWEM is used, by iteratively modifying a candidate dataset in order to make it more consistent with the published marginals.

\subsection{Current State-of-the-art}
  The state of the art (SOTA) approaches are also based on marginal queries. 
  These all follow the ``select-measure-generate''  paradigm, which is arguably as simple as it sounds: given a target workload of queries to answer, select a next marginal that will give the biggest increase in accuracy, measure that marginal (with DP noise), and use the current set of published marginals to generate a set of synthetic data \cite{LiuVW21}. 
The idea was first put forward within the MST paper, and has been adopted by several subsequent works.

The recent AIM method embodies this approach~\cite{McKennaMSM22}. %
It follows the select-measure-generate paradigm using PGM to build a graphical model over noisy marginals. The main contributions are a set of heuristics that greatly improve the utility and scalability of the graphical model, refining the select-measure-generate approach to be more adaptive. These include extending the marginal selection strategy to better account for the signal-to-noise ratio from measuring marginals under DP, a budget annealing strategy which progressively decreases the noise parameters if the model has stopped improving and constraints on the size of the graphical model to prevent it from exploding in memory. 

Alternatives that follow the select-measure-generate paradigm include RAP and its successor RAP++.
In RAP, a collection of queries are posed to the data, and a synthetic dataset is generated by trying to pick one that maximally agrees with the queries~\cite{AydoreBKKM0S21}. This forgoes a graphical model and instead directly models a differentiable continuous representation of the synthetic dataset.
The process is iterated, by (privately) picking a batch of new queries that elicit high error, so that the synthetic dataset can be tuned to better fit them. 
The chief challenge to overcome is in fitting the synthetic data to the query answers.
RAP maintains queries that are differentiable over the data space, allowing powerful optimization methods to be used, before applying rounding to obtain a final output. 
RAP++ extends this approach to handle numeric values in addition to categoric attributes~\cite{VietriAABK0STW22}. 
It introduces additional technical steps to achieve this, namely random linear projections to handle mixtures of numeric and categoric data, and a sigmoid approximation to allow threshold queries to be differentiated.

\subsection{Extensions}
  Marginal-based methods are an active research topic, and several approaches have sought to extend marginal-based tabular data generation in different ways. 
  The PrivLava algorithms aims to support multi-table generation via latent variables~\cite{CaiXC23}; JAM-PGM seeks to handle the case when some of the training data is considered public ~\cite{FuentesMMMS24}; and Private-GSD makes use of genetic algorithms to support numerical features and better fit the training data~\cite{LiuTV023}.

\section{Deep learning-based methods}
\label{sec:dl}
\subsection{Generative Adversarial Networks (GANs)}

In contrast to marginal-based methods, deep-learning approaches leverage neural-networks to directly model synthetic data distributions. Generative Adversarial Networks (GANs) are one such method for producing synthetic data \cite{GoodfellowPMXWOCB14}. The idea is to maintain a generator network $G(z)$ which produces synthetic samples from random noise $z$ and a discriminator network $D(x)$, whose goal is to classify whether a sample is real or fake. These models are trained adversarially, with the generator producing synthetic samples whilst the discriminator attempts to classify it. If the generator can successfully trick the discriminator then it means the GAN can produce sufficiently high-quality synthetic data. 

Adapting GANs to tabular data is not straightforward as tabular data consists of a mix of discrete (categorical) and continuous (numerical) features. To address these challenges, CTGAN was developed specifically for tabular data \cite{XuSCV19}. Xu \etal show how to modify the conventional GAN framework by incorporating techniques to model discrete and continuous variables more effectively. More specifically, continuous columns are modeled using a variational Gaussian mixture model which captures the modes of the distribution and are used for normalizing continuous features. Furthermore, a conditional GAN structure is employed which samples feature values based on its log-frequency and conditionally samples from the GAN using these values. This sampling process allows CTGAN to more evenly explore all possible discrete values to produce more consistent tabular data.

Differentially private variants of GANs also exist. To ensure formal DP guarantees, the GAN is trained via DP-SGD \cite{abadi2016deep}, a general algorithm that can be applied to provide DP guarantees to any neural-network. DP-SGD involves, at each iteration, clipping the individual gradients of the SGD update and adding calibrated Gaussian noise to them, ensuring privacy of the final model. Fang \etal propose DP-CTGAN \cite{fang2022dp}, a DP variant which involves simplifying the mode-based normalization and log-frequency sampling to be more privacy friendly whilst training the model with DP-SGD. %

Despite these advances, GAN-based methods suffer from convergence instability, which can result in inconsistent performance and difficulty in reliably reproducing the underlying source data distributions, including even simple one-way marginals \cite{ganev2024graphical}. Alternatives based on variational autoencoders (VAEs) were also proposed by Xu \etal In particular, the tabular VAE (TVAE) method uses the CTGAN feature modeling framework but explicitly models the data distribution through an encoder-decoder framework. TVAEs helps avoid some of the convergence issues associated with GANs to more reliably produce accurate synthetic data.

\subsection{Extensions to DP-GANs}
While GANs are one solution for generating tabular data they encounter a myriad of problems in practice. In particular, the adversarial training of GANs can cause convergence problems such as mode collapse which is worsened under the noise introduced from differential privacy \cite{thanh2020catastrophic, ganev2024graphical}. Because of this, many private alternatives have been proposed that make use of generator networks but avoid either the adversarial training of GANs or the use of DP-SGD, relying instead on simpler privacy mechanisms.

One of the first extensions to DP-GANs is PATE-GAN~\cite{jordon2018pate} which extends the Private Aggregation of Teacher Ensembles (PATE) framework to GANs. Here the discriminator network $D(x)$ is replaced by PATE which trains a set of teacher discriminators $\{T_i(x)\}_i$ without DP and then releases their predictions under a majority vote via DP. To train the generator network $G(z)$, a student discriminator $S(x)$ is trained on synthetic data generated from $G(z)$ with the labels classified (privately) from the teacher discriminators. This student discriminator is used as a proxy for the discriminator network in a traditional GAN setting and is used to update the generator network. This entire process relies on the post-processing property of differential privacy, since the teacher models are the only ones to access the underlying private data and only their predictions are privatized. Thus the student models and the updates to the generator network remain differentially private via post-processing. Jordon \etal show PATE-GAN always outperforms DP-GAN, likely because PATE-GAN avoids the use of DP-SGD which causes traditional DP-GANs to suffer large losses in utility.

Further methods also replace the discriminator network with more privacy-friendly alternatives. Harder \etal propose DP-MERF \cite{HarderAP21} which aims to simplify the GAN-based training procedure. Instead of using a discriminator network, the idea is to train a generator network on random Fourier feature representations of kernel mean embeddings with the goal of minimizing the Maximum Mean Discrepancy (MMD). This simpler approach is better suited for differential privacy as noise is applied to only the mean embeddings and the generator network is trained on these as a form of post-processing. DP-MERF is shown to consistently outperform other DP-GAN variants. 

Some methods look to combine the marginal-based framework with generator networks. Liu \etal propose GEM \cite{LiuVW21}, a marginal-based method that follows the standard \say{select-measure-generate} paradigm like AIM or MST. However, instead of using a graphical model, it trains a generator network on the noisy marginals that are measured. This hybrid method combines the flexibility and privacy-friendly approach of workload-based marginal methods with the modeling power of generator networks.

\subsection{Recent Advances: Diffusion-based Models and LLMs}
\label{sec:diffusion}

Diffusion models offer a compelling alternative to GANs by addressing several of their limitations. Unlike GANs that rely on adversarial training, which often leads to instability, diffusion models operate by gradually perturbing data via statistical noise, learning a model to reverse this process to recover the original data \cite{sohl2015deep, ho2020denoising}. 
The denoising framework results in a more stable training process and the reverse denoising process can be applied to random noise to generate synthetic data. 
For image generation, recent diffusion models are more effective than GAN-based models \cite{dhariwal2021diffusion}. 

TabDDPM \cite{KotelnikovBRB23} adapts diffusion models to tabular synthetic data. The difficulty in extending diffusion models to tabular data is to adapt the (de)noising process to account for both numerical and categorical features. TabDDPM applies the diffusion noise process independently to each feature. For numerical features, a quantile transformation is applied and a standard Gaussian diffusion model is used. For categorical features, a one-hot encoding is applied and a multinomial noise diffusion process is used. TabDDPM is shown to be far more effective than CTGAN and TVAE for retaining distributional statistics over both categorical and numerical features.

One limitation of TabDDPM is that it considers a diffusion process which adds noise to each feature independently. Zhang \etal propose TabSyn \cite{zhang2023mixed} building directly on TabDDPM to address this issue. It first embeds tabular data into a continuous latent space using a variational autoencoder (VAE) with transformer-based encoders and decoders, unifying both numerical and categorical features into a single latent space. It then applies a standard Gaussian diffusion model to these latent representations to generate synthetic samples. This approach combines both numerical and categorical features into a unified representation and circumvents the challenge of modeling mixed-type tabular data. This modeling approach outperforms TabDDPM on a wide-range of tasks, highlighting the effectiveness of using a latent representation.

Recent approaches have moved away from diffusion-based models to focus instead on leveraging the effectiveness of large-language models (LLMs). Borisov \etal propose GReaT~\cite{borisov2022language} which fine-tunes a pretrained LLM on textually encoded tabular data along with an effective sampling scheme to generate synthetic tabular data from the LLM. This sampling scheme relies on the fact that LLMs are auto-regressive models that generate tokens based on the prior tokens that have been sampled before and a similar procedure can be used to generate the feature values for synthetic tabular data. This process is shown to beat CTGAN and TVAE on a range of tabular benchmarks.

Differentially private adaptations also exist for LLMs. Sablayrolles \etal propose SynLM~\cite{sablayrolles2023privately} which trains a transformer-based language model from scratch via DP-SGD, providing formal privacy guarantees. One difficulty when using LLMs is that due to their auto-regressive nature they are prone to generating invalid values for features. SynLM utilizes a trie structure during generation to ensure that sampling from the LLM produces only valid tokens (i.e., valid feature values) for the synthetic tabular data. Experiments show that SynLM has competitive performance with SOTA marginal-based methods like AIM.

\subsection{Deep-learning vs. Marginal-based methods}

There has been much debate about whether deep-learning approaches outperform marginal-based methods when using differential privacy. Whilst deep generative models offer greater flexibility and may be more naturally suited for learning complex interactions in high-dimensional data, it has often been shown that the simpler marginal-based approaches remain more effective, especially due to the noise added from DP \cite{liu2022utility, tao2021benchmarking, ganev2024graphical, chen2025benchmarking}.

Tao \etal present one such systematic benchmark of various DP-SDG algorithms for tabular data, studying 12 recent DP-SDG methods across 7 benchmark datasets \cite{tao2021benchmarking}. They compare GAN-based approaches such as DP-CTGAN and PATE-GAN against marginal-based methods like MST and MWEM-PGM.  Their experiments reveal that although no one DP-SDG algorithm dominates, the majority of GAN-based methods struggle to accurately reproduce basic statistical properties of the reference dataset such as one-way marginals. This creates synthetic datasets that fail to capture the very basic distributional characteristics of the underlying source data. On the other hand, marginal-based methods consistently show robust performance across both simple statistics and ML tasks, outperforming all GAN-based competitors.

Liu \etal study this phenomena more closely, investigating what they coin the \say{utility recovery incapability} of neural-network based methods \cite{liu2022utility}. Specifically, they highlight that increasing the privacy budget (i.e., relaxing privacy constraints) does not necessarily lead to improved utility of synthetic data for neural-network methods. Their experiments reveal that DP-CTGAN’s performance remains stagnant regardless of the privacy budget, while PATE-GAN only shows utility improvements when the privacy budget is sufficiently large (e.g., $\varepsilon > 3$). They observe marginal-based approaches like PrivBayes do not suffer from these issues, suggesting marginal-based methods may inherently be more stable in terms of utility recovery when increasing the privacy budget.

Ganev \etal compare graphical models with deep generative models for DP synthetic data generation \cite{ganev2024graphical}. Specifically, they focus on how privacy budgets are allocated. Graphical models like PrivBayes and MST distribute budget per column, while deep generative models (e.g., DP-CTGAN and PATE-GAN) spend privacy budget per training iteration. Their findings indicate that graphical models are particularly effective for datasets with limited features and simple tasks, as they preserve basic statistical properties more reliably. However, deep generative models exhibit greater flexibility when handling high-dimensional data, noting that PATE-GAN remains competitive on high-dimensional tasks. Despite this potential, the performance of neural-network approaches can be unpredictable, especially under strict privacy levels ($\varepsilon < 1$).
While recent studies refute the usefulness of private deep-learning methods for tabular synthetic data \cite{chen2025benchmarking}, there still remains a gap where SOTA approaches based on diffusion models and LLMs have not been systematically compared to marginal-based methods. Indeed, Ganev \etal hint at the potential of deep-learning approaches for outperforming marginal-based methods when applied to high-dimensional tabular data, and it may be the case that recent advances can close this gap.

\section{Attacks and defenses}
\label{sec:mia}
Privacy attacks can assess information leakage in procedures using personal data. 
These can  \textit{(a)} quantify privacy leakage for procedures with a-priori unclear or unquantified privacy-preserving properties, and/or \textit{(b)}  provide more ``realistic'' privacy leakage as opposed to worst-case guarantees such as (theoretical) DP.

\subsection{Setup}
An empirical privacy measurement is modeled as a game between an attacker (``the adversary'') and the data curator, described via the following dimensions. We will elaborate on each of the below.

\begin{itemize}
    \item \textbf{Threat Model}: Assumptions on the attacker.
    \item \textbf{Privacy Risk}: Notion of privacy leakage being measured.
\end{itemize}

\subsubsection{Threat Model}
This consists of the following.

\paragraph{Model Access.} 
To what extent is the adversary able to leverage insight into the process of synthetic data modelling?  Options include,
\begin{itemize}
    \item \textbf{Published/No-box}: Access to synthetic data only.
    \item  \textbf{Blackbox}: Sampling access to the synthetic data generator.
    \item \textbf{Whitebox}: Model parameters of the generator known.
     \item \textbf{Active whitebox:} Access to internal states (useful in auditing DP) and ability to modify internal states (of the model).
\end{itemize}

\paragraph{Auxiliary information.}  Auxiliary data is used to model additional relevant information an adversary may have. The strongest adversary, which corresponds to the threat model with respect to which DP is defined, knows all the training (or reference) points except presence/absence of the point which we are testing for membership inference.
\begin{itemize}
    \item \textbf{Aux-train}: The part of the training data that is accessible to the adversary.
    \item \textbf{Aux-test}: The part of the test data that is accessible to the adversary (or access to distribution).
    \item \textbf{Aux}: The accessible part of the training and test data.
\end{itemize}
\paragraph{
Power of Attacker.} This is mostly useful for auditing the DP guarantees by considering the worst-case bounds
\begin{itemize}
    \item \textbf{Active target}: Attacker can choose the point to attack.
    \item \textbf{Active training data:} Attacker chooses the training dataset.
\end{itemize}

\subsubsection{Privacy Risk}
Various notions of privacy risks have been studied in the literature. We elaborate on a few.

\paragraph{Membership Inference.} 
Given a synthetic data generation algorithm and a target data point (and auxiliary information), the goal is to infer presence/absence of the given target data point in the underlying training dataset. Given a membership inference classifier/attack, we can calculate its \textit{True Positive rate (TPR)} and \textit{False Positive rate (FPR)}. 
We obtain an empirical value of the differential privacy parameter $\varepsilon$ for a corresponding $\delta$ as follows \cite{kairouz2015composition}.
\begin{align}
    \varepsilon \geq \log\left({\max\left(\frac{TPR-\delta}{FPR}, \frac{TNR-\delta}{FNR}\right)}\right)
\end{align}

\paragraph{Attribute inference.} Attribute inference is another notion wherein the goal is, given a partial description of a data point used in the training set, to infer its sensitive attributes.  
Besides, other related notions such as those of linkability, singling-out, etc. have also been studied in the literature \cite{lautrup2025syntheval}.

\subsection{Methods and Techniques for Privacy Attacks}

We limit our discussion to membership and attribute inference problem.
Most of the prior works can be broadly divided into two categories, applying to both the above targets.

\paragraph{Density-based attacks.}
A line of work proposed attacks based on the rationale that a synthetic data generation model produces samples ``close" to its training set, thus overfitting to it. 
This leads to a strategy to test based on an estimate of the likelihood of the target point, under the synthetic data generating distribution. 
The work of Hillprecht \etal\cite{hilprecht2019monte} estimates it via Monte-Carlo integration, measuring the fraction of points produced within a neighborhood of the target, under an appropriate distance metric. Subsequently, the works of Hayes \etal\cite{hayes2017logan}
and Van Breugel \etal\cite{van2023membership} instead estimates it by fitting a generative model on the synthetic (and auxiliary) data. The attack, DOMIAS \cite{van2023membership}, operates as follows:
\begin{enumerate}
    \item Fit generative models on synthetic dataset $G$, and auxiliary dataset $R$.
    \item Do a likelihood ratio test on the target, as follows.
    \begin{align}
    A(x_\text{target}) = f\left(\frac{p_G(x_\text{target})}{p_R(x_\text{target})}\right)
\end{align}
\end{enumerate}

where $f$ is a monotonically increasing function with range $[0,1]$. The normalization, using the auxiliary data above, can be interpreted as calibrating the attack \cite{watson2021importance}.
The work of \cite{golob2024privacy}, proposes MAMA-MIA, which instantiates the above framework with a density estimator tailored to marginal-based synthetic data algorithms (such as AIM). The idea therein is to use the relative frequency of the target, with respect to the marginals selected by the algorithm,  as the density estimator. 

\paragraph{Shadow modeling-based attacks.}
A line of work on privacy attacks in supervised learning \cite{shokri2017membership, carlini2022membership, ye2022enhanced, zarifzadeh2023low} relies on what is called \textit{shadow modeling}. The idea is to cast the membership inference attack problem as a supervised learning problem. This approach creates labeled shadow datasets for the membership inference problem by sampling from the auxiliary dataset. A meta-classifier (or a hypothesis test) is trained on the features extracted from the shadow datasets (via synthetic data generation, in our case), to do membership inference on the target.
The approach \cite{stadler2022synthetic,houssiau2022tapas, annamalai2024you} is:
\begin{enumerate}
    \item Generate multiple shadow datasets (from aux-test), by random sampling.
    \item Fit a synthetic data generation model on these datasets, and sample to generate synthetic datasets.
    \item Extract features by querying the datasets (simple statistics, histograms, and correlations).
    \item 
Use these (features, label), as a new dataset, and train a classifier to predict membership inference.
\end{enumerate}

The above approach, due to Annamalai \etal \cite{annamalai2024you}, was able to find privacy vulnerabilities in implementations of many popular synthetic data algorithms, such as PrivBayes \cite{zhang2017privbayes} from DataSynthesizer \cite{ping2017datasynthesizer} and MST \cite{mckenna2021winning} from SmartNoise \cite{kopp2021microsoft}.

\section{Advanced Topics and Open Problems}
\label{sec:open}

\paragraph{Graph Data.}
Graphs are a natural structured way to represent network-based data.  
Graphs can of course be represented as tables, inasmuch as that we can think of an adjacency matrix as a table with binary entries. 
However, it should be clear that methods designed for tabular data generation are not appropriate for graph generation. 
In particular, the semantics of columns of the adjacency matrix (representing the pattern of connections to some node $i$), do not tally with the idea of a column representing a feature in a table. 
In a graph, we have $d=n$, whereas for tables we expect $d \ll n$. 
Instead, synthetic graphs can be generated by specific models targeting this task. 
One line of work seeks to capture graph behaviors based on a few high-level features, such as degree distribution, or the prevalence of triangles in the reference data. 
A generative model of a graph is obtained by randomly sampling from the space of graphs that share these parameters~\cite{Albert:Barabasi:01}. 
Another direction starts with a reference graph, and applies perturbations such as edge addition/removal to mask the ground truth. 
In between are efforts that seek produce richer generative models that are both faithful to the reference data, and providing stronger privacy guarantees~\cite{TRPP22}. 

\paragraph{Image and Video Data.}
The recent advances in generative AI have opened the door to synthetic images and videos, generated based on (text) prompts~\cite{RenAIssance}. 
The complex real-world semantics of visual data, along with the lack of simple structure, mean there is little direct interplay between this modality and tabular data. 
However, as noted in Section~\ref{sec:diffusion}, there are some efforts to apply ideas from diffusion models to tabular data. 

\paragraph{Text Data.}
Synthetic text can be viewed as another instance of generative AI. 
The standard application of Large Language Models (LLMs) is, given some initial text, to generate synthetic text that is a plausible continuation, based on the patterns learned from the (massive) training data~\cite{LLMoverview}. 
The highly structured approach of tabular generation seeks to minimize the opportunity for creativity in the response, and instead to more faithfully provide output that closely matches patterns from the (comparatively tiny) reference data. 
Nevertheless (see Section~\ref{sec:diffusion}), there are efforts to generate tabular data leveraging LLMs~\cite{sablayrolles2023privately,borisov2022language}, and emerging efforts to fuse these approaches to provide more constrained synthetic text.

\paragraph{Distributed data generation.}
The standard model for synthetic data generation assumes that the reference dataset is held centrally by a trusted entity who can access it freely in order to build a generative model. 
However, it is increasingly common to consider scenarios where the data is sharded across a large collection of participants, who wish to cooperate to build the model but who cannot simply pool their data. 
In the machine learning community, this is understood as distributed ML or Federated Learning~\cite{flsurvey}.
Several recent works have studied what it means to perform synthetic data generation with distributed training data. 
The first approaches take a state-of-the-art method from the centralized setting, and study how to implement it over distributed data. 
The tradeoffs here are how to avoid excessive communication overhead. 
This can be done by choosing an algorithm which does not require strict synchronization, or by relaxing the algorithm to allow participants to proceed independently of each other between occasional synchronization steps. 
Pereira \etal~\cite{pereira2022secure} adapt the MWEM algorithm under the model of secure multi-party computation (MPC), leveraging the fact that the relatively simplicity of MWEM can be expressed using lightweight MPC steps. 
Pentyala \etal~\cite{pentyala2024caps} follow a similar approach, expanding to cover PGM-based approaches such as AIM. 
In parallel, Maddock \etal~\cite{maddock2024flaim} seek to emulate the AIM algorithm under the \textit{federated} model, using only simple secure addition primitives to securely collect aggregated (and differentially private) information from distributed clients.  
This reduces the communication and synchronization overhead, but yields weaker accuracy than is achieved in the centralized or fully distributed settings. 

\paragraph{Pragmatic data generation.} 
A common assumption with differential privacy is that the entire dataset is assumed to be private. Recent work has studied public-assisted methods which utilize public data to improve utility \cite{bie2022private, ullah2024public}. 
For synthetic data, it is common in practice that only a subset of the reference data will be privacy-sensitive. 
This may be due to access to a publicly available dataset or that some subset of columns is assumed public-knowledge i.e., age statistics can be obtained from public census information. Leveraging this public information effectively can help improve the utility of the generated synthetic data.  
Current public-assisted synthetic data methods have extended marginal-based approaches to study both a horizontal and vertical public-private partitioning. 
In the horizontal setting, it is assumed that a subset of rows from the reference dataset (following the same schema) is publicly available, whereas in the vertical setting a subset of columns are instead assumed to be public. Common approaches rely on pretraining the synthetic data model on the public dataset. These include PMW-Pub \cite{liu2021leveraging}, a public-assisted version of the MWEM algorithm and GEMPub~\cite{LiuVW21} a public-assisted version of GEM. In both cases, the methods pretrain the synthetic model (multiplicative weights for PMW-Pub and a generator network for GEMPub) on public data before preceding with the standard \say{select-measure-generate} paradigm. Current SOTA methods include JAM-PGM~\cite{FuentesMMMS24}, which uses public information during the private training of AIM, extending the selection strategy to choose between a publicly available marginal or a private one and Conditional AIM \cite{maddock2025leveraging}, which leverages conditional generation in the vertical setting.

\paragraph{Synthetic Data Frameworks.} Many software libraries exist for synthetic tabular data generation, from mature research frameworks \cite{SDV, synthcity} to commercial solutions \cite{gretel, mostlyai}. However, it's common for existing libraries to not include current SOTA methods or focus on specific SDG methods (i.e., deep-learning based only). We plan to open-source our internally developed synthetic data generation library in the future.

\begin{acks}
The work of GC is supported in part by EPSRC grants EP/V044621/1 and UKRI Prosperity Partnership Scheme (FAIR) EP/V056883/1.
\end{acks}

\clearpage

\bibliographystyle{abbrv}
\balance
{\bibliography{tutorial}}

\end{document}